\documentclass[english]{article}
\usepackage{geometry}
\usepackage{amsmath}
\usepackage{float}
\usepackage{xurl}
\usepackage{amsmath,amsfonts,amsthm,bm}
\usepackage{mathtools}
\usepackage{csquotes}
\usepackage{graphicx}
\usepackage{multirow}
\usepackage{amssymb}
\usepackage{wrapfig}
\usepackage[font=footnotesize,labelformat=simple]{subcaption}

\usepackage{graphicx}
\usepackage{bbold}
\usepackage{colortbl}
\usepackage{wrapfig}
\usepackage[bottom]{footmisc}
\usepackage[font=small,labelfont=bf,justification=centering]{caption}
\usepackage{tabularx}
\usepackage{etoolbox}
\usepackage{adjustbox}
\usepackage{pdflscape}
\usepackage{currency}
\usepackage{lipsum}
\usepackage{enumitem}
\usepackage{wrapfig}
\usepackage{booktabs}
\usepackage{caption}
\usepackage{upgreek}
\usepackage{cite}
    
\begin{document}

Accepted for publication in IEEE Access, vol. 11, pp. 8889-8903, doi:10.1109/ACCESS.2023.3239549.

\title{\Huge Fighting Money Laundering with
Statistics and Machine Learning}
\author{Rasmus Ingemann Tuffveson Jensen$^{*,\dag}$ and Alexandros Iosifidis$^{\dag}$
\vspace{0.25cm}\\
$^{*}$Spar Nord Bank, Denmark \\
$^{\dag}$Department of Electrical and Computer Engineering, Aarhus University, Denmark
}

{\let\newpage\relax\maketitle}

\begin{abstract}
Money laundering is a profound global problem. Nonetheless, there is little scientific literature on statistical and machine learning methods for anti-money laundering. In this paper, we focus on anti-money laundering in banks and provide an introduction and review of the literature. We propose a unifying terminology with two central elements: (i) client risk profiling and (ii) suspicious behavior flagging.  We find that client risk profiling is characterized by diagnostics, i.e., efforts to find and explain risk factors. On the other hand, suspicious behavior flagging is characterized by non-disclosed features and hand-crafted risk indices.  Finally, we discuss directions for future research.  One major challenge is the need for more public data sets. This may potentially be addressed by synthetic data generation.  Other possible research directions include semi-supervised and deep learning, interpretability, and fairness of the results.
\end{abstract}

\section{Introduction} 
\label{sec:introduction}

Officials from the United Nations Office on Drugs and Crime estimate that money laundering amounts to 2.1-4\% of the world economy \cite{Pietschmann2001}. The illicit financial flows help criminals avoid prosecution and undermine public trust in financial institutions \cite{McDowell2001,Ferwerda2013,Bartlett2002}. Multiple intergovernmental and private organizations assert that modern statistical and machine learning methods hold great promise to improve anti-money laundering (AML) operations \cite{Grint2017,FATF2021OPP,WolfsbergGroup,Worldbank2020,McKinsey2017}. The hope, among other things, is to identify new types of money laundering and allow a better prioritization of AML resources. The scientific literature on statistical and machine learning methods for AML, however, remains relatively small and fragmented \cite{Leite2019, Ngai2011, Hilal2022}.

The international framework for AML is based on recommendations by the Financial Action Task Force (FATF) \cite{FATF2021}. Within the framework, any interaction with criminal proceeds practically corresponds to money laundering from a bank perspective (regardless of intent or transaction complexity) \cite{Magnusson2009}. Furthermore, the framework requires that banks:
\begin{enumerate}
\item know the identity of, and money laundering risk associated with, clients, and 
\item monitor and report suspicious behavior. 
\end{enumerate} 
Note that we, to reflect FATF's recommendations, are intentionally vague about what constitutes ''suspicious'' behavior.

To comply with the first requirement, banks ask their clients about identity records and banking habits. This is known as know-your-costumer (KYC) information and is used to construct risk profiles. The profiles are, in turn, often used to determine intervals for ongoing due diligence, i.e., checks on KYC information. 

\begin{figure}[!t]
	\centering
	\includegraphics[width=0.66\textwidth]{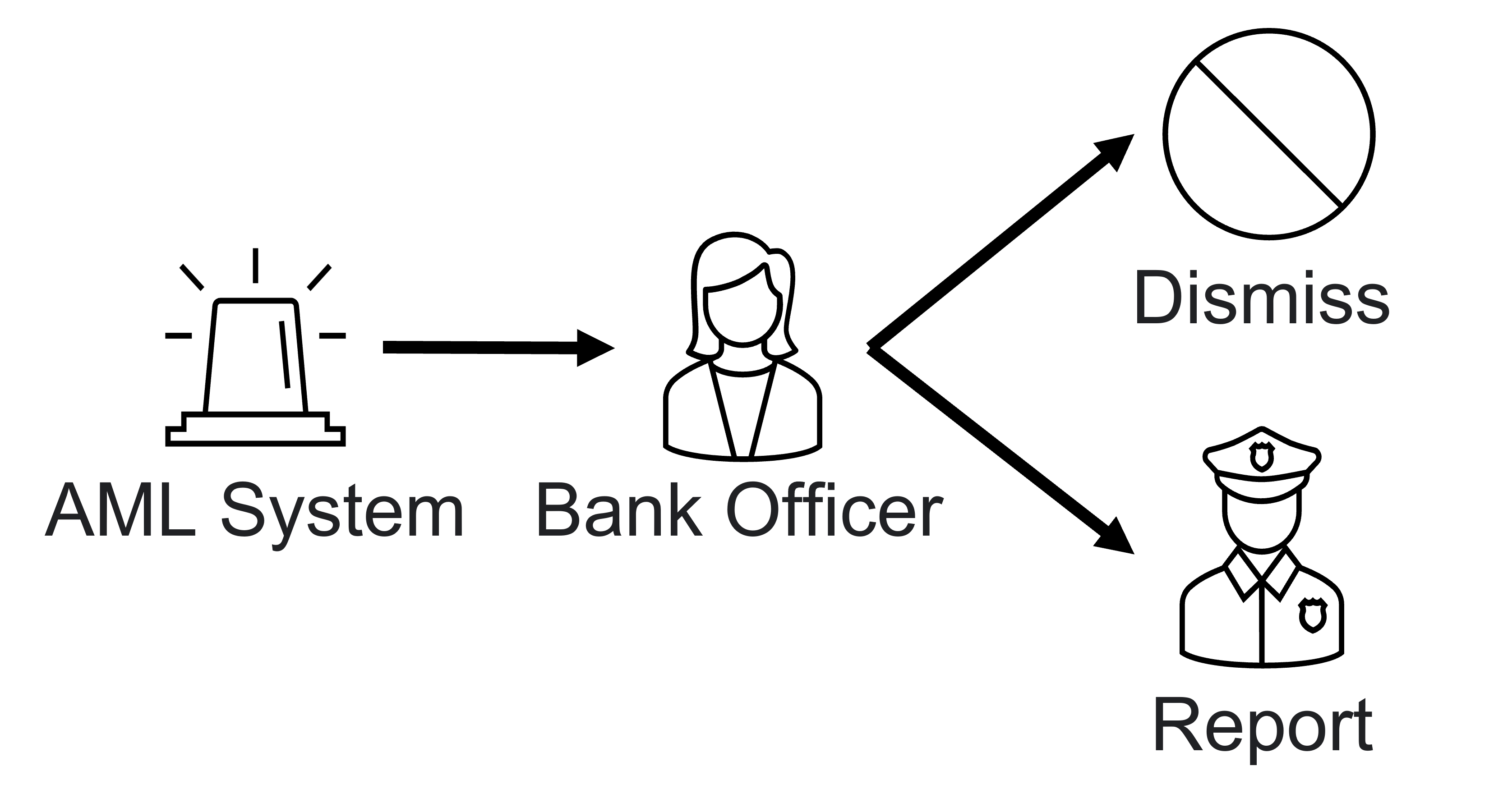}
	\caption{Process of an AML alarm. First, an AML system raises the alarm. A bank officer then reviews it. Finally, it is either dismissed or reported to authorities.}
	\label{fig:alarm process}
\end{figure}

To comply with the second requirement, banks use electronic AML systems to raise alarms for human inquiry. Bank officers then dismiss or report  the  alarms to national financial intelligence units (i.e., authorities). The process is illustrated in Figure \ref{fig:alarm process}.  Traditional AML systems rely on predefined and fixed rules \cite{Verhage2009, Demetis2018}. Although the rules are formulated by experts,  they are essentially `if-this-then-that' statements; easy to interpret but inefficient. Indeed, over 98\% of all AML alarms can be false positives \cite{Richardson2019}.  Banks are not allowed to disclose information about alarms and generally receive little feedback on filled reports. Furthermore, money launderers may change their behavior in response to AML efforts. For instance, banks in the United States must, by law, report all currency transactions over \$10,000 (regardless of whether they constitute money laundering or not) \cite{Sun2021}. In response, money launderers may employ smurfing (i.e., split up large transactions).  Finally, as money laundering has no direct victims, it  can potentially go undetected for longer than other types of financial crime (e.g., credit card or wire fraud).  

In this paper, we focus on AML in banks and aim to provide a technical review that researchers and industry practitioners (statisticians and machine learning engineers) can use as a guide to the current literature on statistical and machine learning methods for AML in banks. Furthermore, we aim to provide a terminology that can facilitate policy discussions, and to provide guidance on open challenges within the literature. To achieve our aims, we (i) propose a unified terminology for AML in banks, (ii) review selected exemplary methods, and (iii) present recent machine learning concepts that may improve AML.

The rest of the paper is organized as follows. Section \ref{sec:Terminology} presents our terminology, distinguishing between (i) client risk profiling and (ii) suspicious behavior flagging. Section \ref{sec:Client Risk Profiling} then reviews the literature on client risk profiling, while Section \ref{sec:Suspicious Behavior Flagging} reviews the literature on suspicious behavior flagging. Note that both Sections \ref{sec:Client Risk Profiling} and \ref{sec:Suspicious Behavior Flagging} contain subsections that further distinguish between unsupervised and supervised methods. Next, Section \ref{sec:Future Research Directions} discusses future research directions. Finally, Section \ref{sec:Conclusion} concludes the paper.

\section{Terminology} \label{sec:Terminology}
Inspired by FATF's recommendations, we argue that banks face two principal data analysis problems in AML: (i) client risk profiling and (ii) suspicious behavior flagging. We use these to structure our terminology and review. A related topic, not discussed here, concerns how authorities treat AML reports (see, for instance, Savage \textit{et al.} \cite{Savage2016}, Drezewski \textit{et al.} \cite{Drezewski2015}, Li \textit{et al.} \cite{Li2017}, or Baltoi \textit{et al.} \cite{Baltoiu2019}).  We further make a distinction between unsupervised and supervised methods. Unsupervised methods  utilize data sets on the form ${\{\mathbf{x}_c \mid c=1,\dots,n\}}$ where $n$ denotes some number of clients. Supervised methods, by contrast, utilize data sets ${\{(\mathbf{x}_c,y_{c}) \mid c=1,\dots,n\}}$ where some labels (e.g., risk scores) $y_c$ are given. 

\subsection{Client Risk Profiling}
Client risk profiling is used to assign general risk scores to clients. Let $\mathbf{x}_c \in \mathbb{R}^d$ be a vector of features specific to client $c$ and $\mathcal{P}$ be a generic set. A client risk profiling is a mapping
\begin{equation}
\rho:\mathbb{R}^d \rightarrow \mathcal{P}, \label{risk profiling}
\end{equation}
 where $\rho(\mathbf{x}_c)$ captures the \textit{money laundering risk} associated with client $c$. For example, we may have ${\mathcal{P}=\{L,M,H\}}$, where $L$ symbolizes low risk, $M$ symbolizes medium risk, and $H$ symbolizes high risk. We stress that client risk profiling in our terminology is characterized by working on the client, not transaction, level. 

\subsection{Suspicious Behavior Flagging}
Suspicious behavior flagging is used to raise alarms on clients, accounts, or transactions. Consider a setup where client $c$ has ${a=1,\dots,A_c}$ accounts. Furthermore, let each account $(c,a)$ have  ${t=1,\dots,T_{(c,a)}}$ transactions and let ${\mathbf{x}_{(c,a,t)} \in \mathbb{R}^d}$ be some features specific to transaction $(c,a,t)$. An AML system is a function
\begin{equation}
s:\mathbb{R}^d \rightarrow \{0,1\},
\end{equation}
where ${s(\mathbf{x}_{(c,a,t)})=1}$ indicates that an alarm is raised on transaction $(c,a,t)$.  Multiple approaches may be used to construct an AML system. Regardless of approach, we argue that all AML systems are  built on one fundamental premise. To cite Bolton and Hand \cite{Bolton2002}: ``\textit{... given that it is too expensive to undertake a detailed investigation of all records, one concentrates investigation on those thought most likely to be fraudulent.}'' Thus, a good AML system needs to model the probability
\begin{equation}
 F(\mathbf{x}_{(c,a,t)}) = P(y_{(c,a,t)} = 1 \mid  \mathbf{x}_{(c,a,t)} ), \label{Prob transaction level}
\end{equation}
where ${y_{(c,a,t)}=1}$ indicates that transaction $(c,a,t)$ should be reported for money laundering (with ${y_{(c,a,t)}=0}$ otherwise). We may then raise alarms given some threshold value ${\epsilon \geq 0}$ and an indicator function \\${s(\mathbf{x}_{(c,a,t)})=1_{\{F(\mathbf{x}_{(c,a,t)}) \geq \epsilon\}}}$.

\begin{figure}[!ht]
	\centering
	\includegraphics[width=0.66\textwidth]{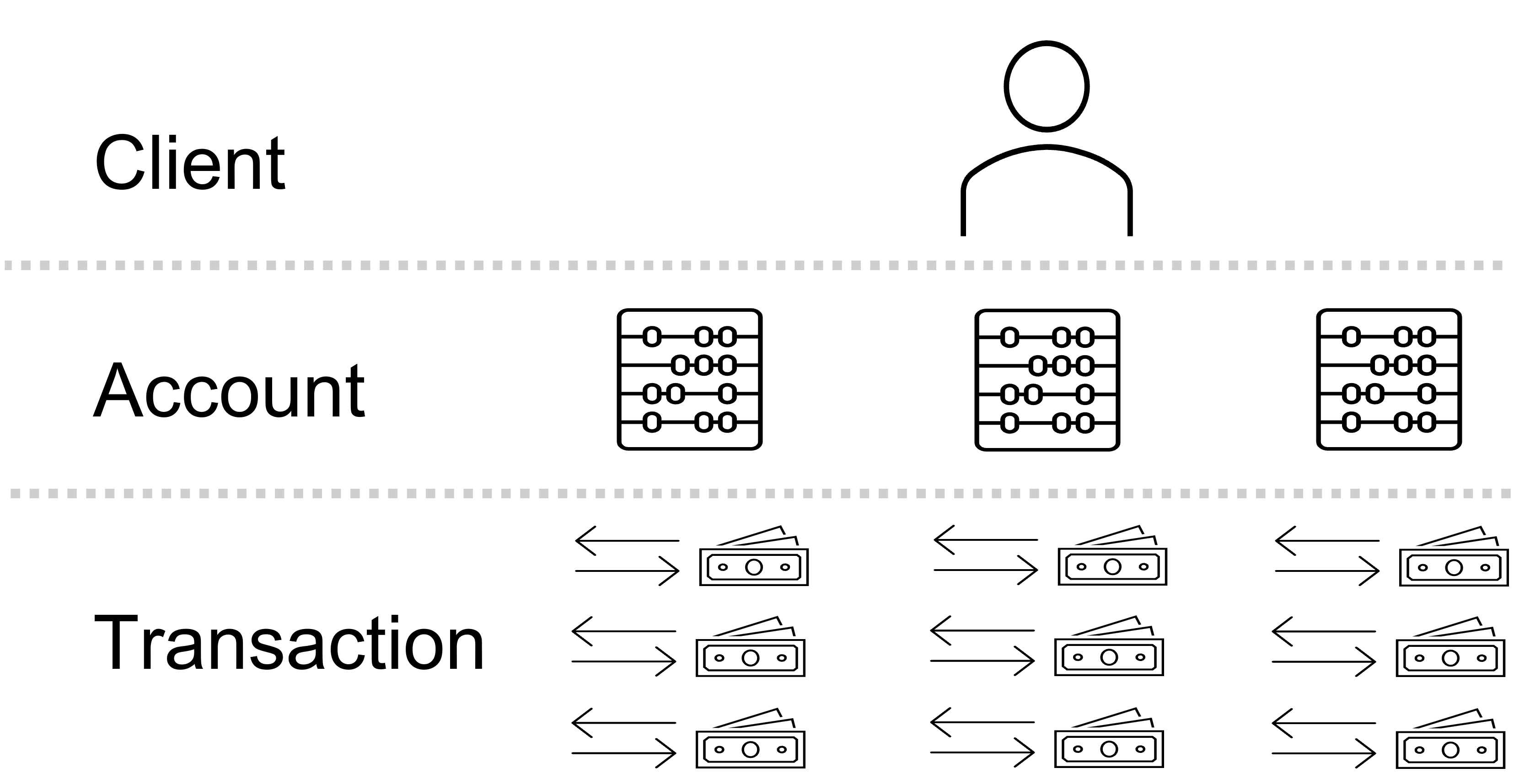}
	\caption{ A single client may hold multiple accounts in a bank, each facilitating numerous transactions. When doing suspicious behavior flagging, alarms may be raised at the client, account, or transaction level (or a combination of them). }
	\label{fig:alarm_levels}
\end{figure}

It can be difficult to determine if a transaction, in itself, is money laundering.  As a remedy, the level of analysis may be changed (see Figure \ref{fig:alarm_levels}). We may, for instance, consider account features ${\mathbf{x}_{(c,a)} \in \mathbb{R}^d}$ that summarize all activity on account $(c,a)$. Alternatively, we may consider the set of all feature vectors ${\mathcal{X}_{(c,a)}=\{\mathbf{x}_{(c,a,1)},\dots,\mathbf{x}_{(c,a,T_{(c,a)})}\}}$ for transactions ${t=1,\dots,T_{(c,a)}}$ made on account $(c,a)$. Defining ${y_{(c,a)}\in\{0,1\}}$ in analogy to $y_{(c,a,t)}$, we may then model
\begin{equation}
 F(\mathbf{x}_{(c,a)}) = P(y_{(c,a)} = 1 \mid  \mathbf{x}_{(c,a)}) \label{Prog account level}
\end{equation} or\begin{equation}
F(\mathcal{X}_{(c,a)}) = P(y_{(c,a)} = 1 \mid  \mathcal{X}_{(c,a)}),
\end{equation}  i.e., the probability that account $(c,a)$ should be reported for money laundering given $\mathbf{x}_{(c,a)}$ or $ \mathcal{X}_{(c,a)}$. Similarly, we could raise alarms directly at the client level, modeling 
\begin{equation}
F(\mathbf{x}_{c}) = P(y_{c} = 1 \mid \mathbf{x}_{c}), \label{Prog client level}
\end{equation}
where $y_{c}\in\{0,1\}$ indicates (with $y_{c}=1$) that client $c$ should be reported for money laundering.  Note that suspicious behavior flagging and client risk profiling  can  overlap at the client level.  Indeed, we could use $F(\mathbf{x}_{c})$ as a risk profile for client $c$.

\section{Client Risk Profiling} \label{sec:Client Risk Profiling}
 We find that studies on client risk profiling are characterized by diagnostics, i.e., efforts to find and explain risk factors. Specifically, unsupervised methods are used to search for new ''risky'' observations or risk factors. On the other hand, supervised methods are used with an explanatory focus. We also find that studies employing unsupervised methods generally use relatively large data sets. By contrast, studies employing supervised methods use small (labeled) data sets. This difference is likely associated with the cost of labeling observations. Finally, we note that while all studies use private data sets, most share a fair amount of information about the features that they use. As we shall see later, this contrasts with the literature on suspicious behavior flagging.

\subsection{Unsupervised Client Risk Profiling} \label{subsec:Unsupervised Client Risk Profiling}
Alexandre and Balsa \cite{Claudio2015} employ $K$-means clustering \cite{Lloyd1982} to construct risk profiles. The algorithm seeks a clustering $\rho:\mathbb{R}^d\rightarrow\{S_1,\dots,S_K\}$ that assigns every client $c$ to a cluster $k=1,\dots,K$. This is achieved by solving for
\begin{equation}
\{\boldsymbol{\mu}_{k}\} = 
\underset{ \{\boldsymbol{\mu}_{k}\} }{\arg\min} \sum_{k=1}^{K}\sum_{\rho_k}\left\Vert \mathbf{x}_{c}-\boldsymbol{\mu}_{k}\right\Vert ^{2}, 
\end{equation}
where $\boldsymbol{\mu}_{k}\in\mathbb{R}^d$ denotes the mean of cluster $k$ and ${\rho_k= \{c=1,\dots,n | \rho(\mathbf{x}_c)=k\}}$ denotes the set of clients assigned to cluster $k$. The problem is addressed in a greedy optimization fashion; iteratively setting $\boldsymbol{\mu}_{k}=\frac{1}{|\rho_k|}\sum_{\rho_k}\mathbf{x}_{c}$ and $\rho(\mathbf{x}_{c})=\arg\min_{k=1,\dots,K}\left\Vert \mathbf{x}_{c}-\boldsymbol{\mu}_{k}\right\Vert ^{2}$. To evaluate the approach, the authors employ a data set with approximately 2.4 million clients from an undisclosed financial institution. Disclosed features include the average size and number of transactions.  The authors implement $K=7$ clusters, designating two of them as risky. The first contains clients with  many  transactions but low transaction values. The second contains clients with older accounts but larger transaction values. Finally, the authors employ decision trees (see Section \ref{subsec:Supervised Client Risk Profiling}) to find classification rules that emulate the clusters. The motivation is, presumably, that bank officers find it easier to work with rules than with $K$-means.

Cao and Do \cite{Cao2012} present a similar study, applying clustering with slope \cite{Yang2002}. Starting with 8,020 transactions from a Vietnamese bank, the authors first change the level of analysis to individual clients. Features include the sum of in- and outgoing transactions, the number of sending and receiving third parties, and the difference between funds  sent  and received. The authors then discretize features and build clusters based on cluster histograms' height-to-width ratios.  They finally simulate 25 accounts with money laundering behavior,  some easily identifiable in the produced clusters. Much may, however, depend on the nature of the simulations. 

Paula \textit{et al.} \cite{Paula2016} use an autoencoder neural network to  find outlier Brazilian export firms. Neural networks are directed, acyclic graphs  connecting  computational units (i.e., neurons) in layers. The output of a feedforward neural network with $l=1,\dots,L$ layers is given by \\ \begin{align} nn&\left(\mathbf{x}_{c}\right)= \nonumber \\ & \phi^{(L)}\Biggl(\biggl(\cdots\Bigl(\phi^{(1)}\bigl(\mathbf{x}_{c}\mathbf{W}^{(1)} +\mathbf{b}^{(1)}\bigl)\Bigl) \cdots\biggl)\mathbf{W}^{(L)}+\mathbf{b}^{(L)}\Biggl),
\end{align} where  $\mathbf{W}^{(1)}\in\mathbb{R}^{d\times h_{1}},\ldots,\mathbf{W}^{(L)}\in\mathbb{R}^{h_{(L-1)}\times h_{L}}$ are weight matrices, $\mathbf{b}^{(1)}\in\mathbb{R}^{h_{1}},\ldots,\mathbf{b}^{(L)}\in\mathbb{R}^{h_{L}}$ are biases, and $\phi^{(1)},\ldots,\phi^{(L)}$ are (non-linear) activation functions. Neural networks are commonly trained with iterative gradient-based optimization.  This includes backpropagation \cite{Rumelhart1986} coupled with stochastic gradient descent \cite{Robbins1951} or more recent adaptive schemes like Adam \cite{Kingma2015}.  The aim is to minimize a loss function $l(o_c,nn(\mathbf{x_c}))$ over all observations $c=1,\dots,n$ where $o_c$ is a target value or vector. Autoencoders, as employed by the authors, are a special type of neural networks that seek a latent representation of their inputs. To this end, they employ an encoder-decoder (i.e., ``hourglass'') architecture and try to replicate their inputs in their outputs, i.e., have $o_c = \mathbf{x}_{c}$. The authors specifically use 5 layers with 18, 6, 3, 6, and 18 neurons. The first two layers (with 18 and 6 neurons) form an encoder.  The middle layer with 3 neurons then obtains a latent representation.  Finally, the last two layers (with 6 and 18 neurons) form a decoder. The approach is tested on a data set with $819,990$ firms. Features include information about debit and credit transactions, export volumes, taxes paid, and previous customs inspections. As a measure of risk, the authors employ the reconstruction error $\rho(\mathbf{x}_{c})=\frac{1}{q}\Vert nn(\mathbf{x}_{c}) - \mathbf{x}_{c} \Vert^2$, frequently used for anomaly or novelty detection in this setting (see, for instance, \cite{Chen2017}). This way, they identify 20 high-risk firms.

\subsection{Supervised Client Risk Profiling} \label{subsec:Supervised Client Risk Profiling}
Colladon and Remondi \cite{Colladon2017} combine social network analysis and logistic regression. Using 33,670 transactions from an Italian factoring firm, the authors first construct three graphs; $\mathcal{G}_{1},\mathcal{G}_{2}$, and $\mathcal{G}_{3}$. All share the same nodes, representing clients, while edges represent transactions. In $\mathcal{G}_{1}$, edges are weighted relative to transaction size. In $\mathcal{G}_{2}$, they are weighted relative to connected clients' business sectors. Finally, in $\mathcal{G}_{3}$, they are weighted relative to geographic factors. Next, a set of graph metrics are used to construct features for every client. These include in-, out-, and total-degrees, closeness, betweenness, and constraint. A label $y_{c}\in\{0,1\}$ is also collected for $288$ clients, denoting (with $y_{c}=1$) if the client can be connected to a money laundering trial. The authors then employ a logistic regression model
\begin{align}
P\left(y_{c}=1\left|\mathbf{x}_{c}\right.\right)=\frac{\exp{(\mathbf{\boldsymbol\beta}^{\intercal}\mathbf{x}_{c})}}{1+\exp{(\boldsymbol\beta^{\intercal}\mathbf{x}_{c})}},
\end{align} where $\boldsymbol{\beta}\in\mathbb{R}^d$ denotes the learnable coefficients. The approach achieves an impressive performance. Results indicate that in-degrees over $\mathcal{G}_{3}$ and total-degrees over $\mathcal{G}_{1}$ are associated with higher risk. By contrast, constraint over $\mathcal{G}_{2}$ and closeness over $\mathcal{G}_{1}$ are associated with lower risk.

Rambharat and Tschirhart \cite{Rambharat2015} use panel data from a financial institution in the United States. The data tracks risk profiles $y_{c_p}\in\{1,2,3,4\}$, assigned to $c=1,\dots,494$ clients over $p=1,..,13$ periods. Specifically,  $y_{c_p}$ represents  low-, medium-, and two types of high-risk profiles. Period-specific features $\mathbf{x}_{c_p}\in\mathbb{R}^d$ include information about clients' business departments, four non-specified ``law enforcement actions'', and dummy (one-hot encoded) variables that capture the time dimension.  To model the data, the authors use an ordinal random effects model where errors and fixed effects are assumed to follow Gaussian distributions. If we let $\Phi(\cdot)$ denote the standard Gaussian cumulative distribution function, the model can be expressed as
\begin{align}
P\bigl(&y_{c_p}=m\left|\mathbf{x}_{c_p},\alpha_{c},\boldsymbol{\beta},\theta_{m},\theta_{m-1},\sigma_{\alpha}\right.\bigl)= \nonumber \\ &\Phi\left(\theta_{m}-\boldsymbol{\beta}^{\intercal}\mathbf{x}_{c_p}-\alpha_{c}\right)-\Phi\left(\theta_{m-1}-\boldsymbol{\beta}^{\intercal}\mathbf{x}_{c_p}-\alpha_{c}\right), 
\end{align}
where $\alpha_{c}$ denotes a random client effect, $\boldsymbol{\beta}\in\mathbb{R}^{q}$ denotes coefficients, and $\theta_{m}$ represents a cut-off value transforming a continuous latent variable $y_{c_p}^{*}$ into $y_{c_p}$.  Specifically, we have $ y_{c_p}=m$ if and only if $\theta_{m-1}<y_{c_p}^* \leq \theta_{m}$. The level of confidentiality makes  it  hard to generalize results from the study. The study does, however, illustrate that banks can benefit from a granular risk rating of  high-risk clients. 

Mart\'inez-S\'anchez \textit{et al.} \cite{Sanchez2020} use decision trees to model clients of a Mexican financial institution. Decision trees \cite{Breiman1984} are flowchart-like models where internal nodes split the feature space into mutually exclusive sub-regions. Final nodes, called leaves, label observations using a voting system. The authors use data on  181 clients, all labeled as either  high-risk or low-risk.  Features include information about seniority, residence, and economic activity. Notably, no train-test split is used. This makes the focus on diagnostics apparent. The authors find that clients with more seniority are comparatively riskier. 

Badal-Valero \textit{et al.} \cite{BadalValero2018} combine Benford's Law and four machine learning models. Benford's Law \cite{Benford1938} gives an empirical distribution of leading digits. The authors use it to extract features from financial statements. Specifically, they consider statements from $335$ suppliers to a company on trial for money laundering. Of these, 23 suppliers have been investigated and labeled as colluders. All other (non-investigated) suppliers are treated as benevolent. The motivating idea is that any colluders, hiding in the non-investigated group, should be misclassified by the employed models. These include a logistic regression, feedforward neural network, decision tree, and random forest. Random forests \cite{Breiman2001}, in particular, combine multiple decision trees. Every tree uses a random subset of features in every node split. To address class imbalance, i.e., the unequal distribution of labels, the authors investigate weighting and synthetic minority oversampling \cite{Chawla2002}. The former weighs observations during training, giving higher importance to data from the minority class. The latter balances the data before training, generating synthetic observations of the minority class. According to the authors, synthetic minority oversampling works the best. However, the conclusion is apparently based on simulated evaluation data.

Gonz\'alez and Val\'asquez \cite{Gonzalez2013} employ a decision tree, feedforward neural network, and Bayesian network to model Chilean firms using false invoices. Bayesian networks \cite{Pearl1985}, in particular, are probabilistic models that represent variable dependencies via directed acyclic graphs. The authors use data on 582,161 firms, 1,692 of which have been labeled as either fraudulent or non-fraudulent. Features include information about previous audits and taxes paid. Because  most  firms are unlabeled, the authors first use unsupervised learning to characterize high-risk behavior. To this end, they employ self-organizing maps \cite{Kohonen2004} and neural gas \cite{Martinetz1991}. Both are neural network techniques that build on competitive learning \cite{Rumelhart1985} rather than  error correction (i.e., gradient-based optimization).  While the methods do produce clusters with some behavioral patterns, they do not appear useful for false invoice detection. On the labeled training data, the feedforward neural network achieves the best performance.

\section{Suspicious Behavior Flagging} \label{sec:Suspicious Behavior Flagging}
 We find that the literature on suspicious behavior flagging is characterized by a large proportion of short and suggestive papers. This includes applications of fuzzy logic \cite{Chen2011}, autoregression \cite{Kannan2017}, and sequence matching \cite{LiuXuan2008}. Very few studies apply outlier or anomaly detection techniques \cite{Hilal2022}. In contrast to work by Canhoto \cite{Canhoto2021}, our review demonstrates that there is ample scope to employ both unsupervised and supervised methods for suspicious behavior flagging. Studies using unsupervised methods, however, often contain little performance evaluation. By contrast, studies that use supervised methods naturally use (a part of) their labeled data for evaluation. In line with thoughts by Breiman \cite{Breiman2002} (on fraud detection), there is some evidence that supervised methods might perform better than unsupervised methods; see the last part of Section \ref{Supervised Suspicious Activity Flagging}. However, different types of employed data and the small size of the literature make it difficult to draw a conclusion. Furthermore, non-disclosed features and hand-crafted risk indices generally make it difficult to compare studies.

\subsection{Unsupervised Suspicious Behavior Flagging}
Larik and Haider \cite{Larik2011} flag transactions with a combination of principal component analysis and $K$-means. Given data on approximately 8.2 million transactions, the authors first seek to cluster clients. To this end, principal component analysis \cite{Jolliffe2002} is applied to client features $\mathbf{x}_c\in \mathbb{R}^d$, $c=1,...,n$. The method seeks lower-dimensional, linear transformations $\mathbf{z}_c\in \mathbb{R}^q$, $q<d$, that preserve the greatest amount of variance. Let $\mathbf{S}$ denote the data covariance matrix. The first coordinate of $\mathbf{z}_c$, called the first principal component, is then given by $\mathbf{u}_{1}^T\mathbf{x}_c$ where the principal direction $\mathbf{u}_{1}\in\mathbb{R}^d$ is determined by \\ 
\begin{equation}
\begin{aligned}
\mathbf{u}_{1} = \underset{\mathbf{u}\in\mathbb{R}^{d}}{\arg \max} \quad & \mathbf{u}^{\intercal} \mathbf{S}\mathbf{u}\\
\textrm{s.t.} \quad & \mathbf{u}^{\intercal}\mathbf{u}=1\\
\end{aligned}
\end{equation} By analogy,  the $j$'th principal component is given by $\mathbf{u}_{j}^T\mathbf{x}_c$ where $\mathbf{u}_{j}\in\mathbb{R}^d$ maximizes $\mathbf{u}_{j}^{T}\mathbf{S}\mathbf{u}_{j}$ subject to $\mathbf{u}_{j}^T\mathbf{u}_{j}= 1$ and orthogonality with the previous principal components $h=1,\dots,j-1$. Principal components are commonly obtained by the eigenvectors of $\mathbf{S}$ corresponding to maximal eigenvalues. Next, the authors use a modified version of $K$-means to cluster $\mathbf{z}_c, \: c=1,\dots,n$. The modification introduces a parameter to control the maximum distance between an observation and the mean of its assigned cluster. A hand-crafted risk index is then used to score and flag incoming transactions. The index compares the sizes and frequencies of transactions within assigned client clusters. As no labels are available, evaluation is limited. 

Rocha-Salazar \textit{et al.} \cite{RochaSalazar2021} mix fuzzy logic, clustering, and principal component analysis to raise alarms. With fuzzy logic \cite{Zadeh1983}, experts first assign risk scores to feature values. These include information about client age, nationality, and transaction statistics. Next, strict competitive learning, fuzzy $C$-means, self-organizing maps, and neural gas are used to build client clusters. The authors find that fuzzy $C$-means \cite{Dunn1973}, in particular, produces the best clusters. This algorithm is similar to $K$-means but uses scores to express degrees of cluster membership rather than hard assignments. The authors further identify one high-risk cluster. Transactions in this cluster are then scored with a hand-crafted risk index. This builds on principal component analysis, weighing features relative to their variances. Data from a Mexican financial institution is used to evaluate the approach. Training is done with 26,751 private and 3,572 business transactions; testing with 1,000 private and 600 business transactions. The approach shows good results on balanced accuracy  (i.e., the average of the true positive and true negative rate). 

Raza and Haider \cite{RAZA2011987} propose a combination of clustering and dynamic Bayesian networks. First, client features $\mathbf{x}_c$ are clustered with fuzzy $C$-means. For each cluster, a $q$-step dynamic Bayesian network \cite{DAGUM199241} is then trained on transaction sequences $\mathcal{X}_{(c,a)}=\{\mathbf{x}_{(c,a,1)},\dots,\mathbf{x}_{(c,a,T_{(a,c)})}\}$.  Transaction features $\mathbf{x}_{c,a,t}$ include information about amount, period, and type.  At test time, incoming transactions (along with the previous $q=1,2$ transactions) are passed through the network. A hand-crafted risk index, building on outputted posterior probabilities, is then calculated. The approach is implemented on a data set with approximately 8.2 million transactions (presumably the same data used by Larik and Haider \cite{Larik2011}). However, as no labels are available, evaluation is limited. 

Camino \textit{et al.} \cite{Camino2017} flag clients with three outlier detection techniques:  an isolation forest, a one-class support vector machine, and a Gaussian mixture model.  Isolation forests \cite{Liu2008} build multiple decision trees using random feature splits. Observations isolated by comparatively few feature splits (averaged over all trees) are then considered outliers. One-class support vector machines \cite{Bernhard2001} use a kernel function to map data into a reproducing Hilbert space. The method then seeks a maximum margin hyperplane that separates data points from the origin. A small number of observations are allowed to violate the hyperplane; these are considered outliers. Finally, Gaussian mixture models \cite{Reynolds2009} assume that all observations are generated by a number of Gaussian distributions. Observations in low-density regions are then considered outliers. The authors combine all three techniques into a single ensemble method. The method is tested on a data set from an AML software company. This contains one million transactions with  client-level features recording summary statistics. The authors report positive feedback from the data-supplying company; otherwise, evaluation is limited. 

Sun \textit{et al.} \cite{Sun_2022} apply extreme value theory \cite{Embrechts1997} to flag outliers in transaction streams. The authors start by engineering two features. The first records the number of times an account has reached a balanced state, i.e., when money transferred into an account is transferred out again. The second records the number of effective fan-ins associated with an account, i.e., when money transferred into the account surpasses a given limit and the account again reaches a balanced state. Next, the Pickands–Balkema–De Haan theorem  \cite{Pickands1975,Balkema1974} is invoked to model (derived) conditional feature exceedances according to a generalized Pareto distribution. The approach allows the authors to flag transactions according to a probabilistic limit $p$  (in analogy to the $p$-values used to test null hypotheses).  The approach is tested on real bank data with simulated noise and outliers.

\subsection{Supervised Suspicious Behavior Flagging} \label{Supervised Suspicious Activity Flagging}
Deng \textit{et al.} \cite{Deng2009} combine logistic regression, stochastic approximation, and sequential $D$-optimal design for active learning. The question is how we sequentially should select new observations for inquiry (revealing $y_{(c,a)}$) and  use them in  the estimation of
\begin{equation}
F(\mathbf{x}_{(c,a)}) = P(y_{(c,a)} = 1 \mid  \mathbf{x}_{(c,a)}).
\end{equation}
 The authors employ a data set with 92 inquired accounts and two highly engineered features. The first feature $x_{(c,a)}^{(1)}\in\mathbb{R}$ captures the velocity and size of transactions; the second $x_{(c,a)}^{(2)}\in\mathbb{R}$ captures peer comparisons. Assuming that $F(\cdot)$ is an increasing function in both features, the authors further define a synthetic variable $z_{(c,a)}=\omega x_{(c,a)}^{(1)}+(1-\omega)x_{(c,a)}^{(2)}$ for $\omega\in[0,1]$. Finally, $z_{(c,a)}$ is subject to a univariate logistic regression on $y_{(c,a)}$. This allows a combination of stochastic approximation \cite{Wu1985} and sequential $D$-optimal design \cite{Neyer1994} for new observation selection. The approach significantly outperforms random selection. Furthermore, simulations show that it is robust to underlying data distributions.

Borrajo \textit{et al.} \cite{Borrajo2020} argue that AML models may benefit from other types of information than simple transaction statistics. To this end, the authors consider behavior trances. These, among other things, contain information about account creation and company ownership. Using custom distance functions, the authors apply  $K$-nearest neighbors  \cite{Fix_Hodges1989,Altman1992} to flag illicit behavior. The method predicts that a new observation belongs to the same class as the majority of its $k$ nearest neighbors. While the authors report excellent results, these are, notably, obtained on simulated data.

Zhang and Trubey \cite{Zhang2019} employ six machine learning models to predict the outcome of AML alarm inquiries. We note that the setup can be used both to qualify existing and raise new alarms under appropriate assumptions. Indeed, let $s_c\in\{0,1\}$ indicate (with $s_c=1)$ that client $c$ is flagged by a traditional AML system. Assuming that $s_c$ and $y_c$ are conditionally independent given $\mathbf{x}_c$, we have that
\begin{equation}
P(y_c=1|\mathbf{x}_c)=P(y_c=1|\mathbf{x}_c,s_c=1).
\end{equation}
If we also assume that $P(s_c=1|\mathbf{x}_c)>0$ for all $\mathbf{x}_c\in \{\mathbf{x}\in\mathbb{R}^d:P(y_c=1|\mathbf{x})>0\}$, we can use a model, only trained on previously flagged clients, to raise new alarms. The authors use a data set with 6,113 alarms from a financial institution in the United States. Of these, 34 alarms were reported to authorities. The data set contains ten non-disclosed features. In order to address class imbalance, the authors investigate random over- and undersampling. Both techniques, in particular, increase the performance of a support vector machine \cite{Cortes1995}. This model seeks to maximize the margin between feature observations of the two classes and a class separating hyperplane (possibly in transformed space). However, a feedforward neural network, robust to both sampling techniques, shows the best performance. 

Jullum \textit{et al.} \cite{Jullum2020} use gradient boosted trees to model AML alarms. The approach additively combines $f_1,\dots,f_K$ regression trees (i.e., decision trees with continuous outputs) and is implemented with XGBoost \cite{Chen2016}. Data comes from a Norwegian bank and contains:
\begin{enumerate}
    \item 16,192 non-flagged transactions,
    \item 14,932 flagged transactions, dismissed after brief inquiries,
    \item 1260 flagged transactions, thoroughly inquired but dismissed, and
    \item 750 flagged and reported transactions.
\end{enumerate}
The authors primarily perform binary classification. Here, transactions in (1)-(3) are treated as licit; transactions in (4) are treated as illicit. Features include information about client background, behavior, and previous AML alarms. To compare model performance with traditional AML systems, the authors propose an evaluation metric called ``proportion of positive predictions'' (PPP). This records the proportion of positive predictions when classification thresholds are adjusted to obtain a pre-specified true positive rate. Results, in particular, indicate that  the  inclusion of type (1) transactions improve performance.  

Tertychnyi \textit{et al.} \cite{Tertychnyi2020} propose a two-layer approach to flag suspicious clients. In the first layer, a logistic regression is used to filter out clients with transaction patterns that are clearly non-illicit. In the second layer,  the  remaining clients are subject to gradient boosted trees implemented with CatBoost \cite{Prokhorenkova2018}. The authors employ a data set from an undisclosed bank. This contains approximately 330,000 clients from three countries.  About 0.004\% of the clients have been reported for money laundering. The remaining are randomly sampled. Client-level features  include demographic data and transaction statistics. Model performance varies significantly over the three countries in the authors' data set. However, the performance decreases when each country is modeled separately.  

Eddin \textit{et al.} \cite{Eddin2021} investigate how aggregated transaction statistics and different graph features can be used to flag suspicious bank client behavior. To this end, the authors consider a random forest model, generalized linear model \cite{Nelder1972}, and gradient boosted trees with LightGBM \cite{Guolin2017}. The authors utilize a large data set from a non-disclosed bank. This contains 500,000 flagged transactions distributed over 400,000 accounts (3\% of which are deemed truly suspicious and labeled as positives). To construct graph features on the data, the authors treat accounts as nodes and transactions as directed edges.  Results indicate that the inclusion of GuiltyWalker features \cite{Oliveira2021}, using random walks to capture the distances between a given node and illicit nodes, increases model performance. 

Charitou \textit{et al.} \cite{Charitou2020} combine a sparse  autoencoder and a generative adversarial network to flag money laundering in online gambling. The sparse  autoencoder  is first used to obtain higher-dimensional latent feature encodings. The goal is to increase the distance between positive (i.e., illicit) and negative (i.e., licit) observations. The latent encodings are then used to train a generative adversarial network \cite{Goodfellow2014}. This is composed of two competing networks. A generative network produces synthetic observations from Gaussian noise. A discriminative network tries to separate these from real observations and determine the class of the observations. The approach is tested on multiple data sets. In an AML context, the most relevant of these pertains to money laundering in online gambling. This data set contains 4,700 observations (1,200 of which were flagged for potential money laundering). 
fi
Weber \textit{et al.} \cite{Weber2019} use graph convolutional neural networks to flag suspicious bitcoin transactions. An open data set is provided by Elliptic, a private cryptocurrency analytics company. The data set contains a transaction graph $\mathcal{G}=(\mathcal{V},\mathcal{E})$ with $|\mathcal{V}|=203,769$ nodes and $|\mathcal{E}|=234,355$ edges. Nodes represent bitcoin  transactions, while edges represent directed payment flows.   Using a heuristic approach, 21\% of the nodes are labeled as licit; 2\% as illicit. For all nodes, 166 features are recorded. Of these, 94 record local   information, while the remaining 72 record one-hop information.   Graph convolutional neural networks \cite{Kipf2016} are neural networks designed to work on graph data. Let $\mathbf{\hat{A}}$ denote the normalized adjacency matrix of graph $\mathcal{G}$. The output of the network's $l$'th layer is obtained by 
\begin{equation} \mathbf{H}^{(l)}=\phi^{(l)}\left(\mathbf{\hat{A}}\mathbf{H}^{(l-1)}\mathbf{W}^{(l)}\right), \end{equation}
where $\mathbf{W}^{(l)}\in\mathbb{R}^{h_{(l-1)}\times h_{(l)}}$ is a weight matrix, $\mathbf{H}^{(l-1)}\in\mathbb{R}^{|V|\times h_{(l-1)}}$ is the output from layer $l-1$ (initiated with feature values), and $\phi^{(l)}$ is an activation function. While the best performance is achieved by a random forest model, the graph convolutional neural network proved competitive. Utilizing a time dimension in the data, the authors also fit a temporal graph convolutional neural network \cite{Pareja2019}. This outperforms the simple graph convolutional neural network. However, it still falls short of the random forest model.

We finally highlight three recent studies that use the Elliptic data set \cite{Weber2019}. Alarab \textit{et al.} \cite{Alarab2020} propose a neural network structure where graph convolutional embeddings are concatenated with linear embeddings of the original features. This increases model performance significantly. Vassallo \textit{et al.} \cite{Vassallo2021} investigate the use of gradient boosting on the Elliptic data. Results, in particular, indicate that gradient boosted trees outperform random forests. Furthermore, the authors propose an adapted version of XGBoost to reduce the impact of concept drift. Lorenz \textit{et al.} \cite{Lorenz2020} experiment with unsupervised anomaly detection. The authors try seven different techniques: local outlier factor \cite{Breunig2000}, $K$-nearest neighbors \cite{Fix_Hodges1989,Altman1992}, principal component analysis \cite{Jolliffe2002}, one-class support vector machine \cite{Bernhard2001} , cluster-based outlier factor \cite{Jiang2008}, angle-based outlier detection \cite{Kriegel2008}, and isolation forest \cite{Liu2008}.  For evaluation, the F1-score is used, recording the harmonic mean between precision and recall (i.e., the true positive rate). Strikingly, all seven unsupervised methods perform substantially worse than a supervised random forest benchmark.  As noted by the authors, this contradicts previous literature on unsupervised behavior flagging (see, for example, \cite{Camino2017}). One possible explanation is that the Elliptic data, constructed over bitcoin transactions, is qualitatively different from bank transaction data. The authors, following Deng \textit{et al.} \cite{Deng2009}, further experiment with four active learning strategies combined with a random forest, gradient boosted trees, and logistic regression model. Two of the active learning strategies build on unsupervised techniques: elliptic envelope \cite{Rousseeuw1999} and isolation forest \cite{Liu2008}. The remaining two build on supervised techniques: uncertainty sampling \cite{Lewis1994} and expected model change \cite{Settles2008}. Results show that the supervised techniques perform the best.

\section{Future Research Directions} \label{sec:Future Research Directions}
 Our review reveals that class imbalance and the lack of publicly available data sets are central challenges to AML research.  Both may motivate the use of synthetic data. We also note how banks hold vast amounts of high-dimensional and unlabeled data \cite{Sudjianto2010}. This may motivate the use of dimension reduction and semi-supervised learning techniques. Other possible research directions include data visualization,  deep learning,  and interpretable and fair machine learning. In the following, we introduce each of these topics. We also provide brief descriptions of related methods and techniques within each topic.

\subsection{Class Imbalance, Evaluation Metrics, and Synthetic Data}
Due to class imbalance, AML systems tend to label all observations as benevolent. This implies that accuracy is a poor evaluation metric.  Instead, we highlight the receiver operating characteristic (ROC) curve \cite{FAWCETT2006861}, plotting true positive versus false positive rates for varying classification thresholds. The area under a ROC curve, called ROCAUC (or sometimes just AUC), is a measure of separability; equal to 1 for perfect classifiers and 0.5 for naive classifiers. Another possible evaluation tool is the precision-recall (PR) \cite{Saito2015} curve, plotting precision versus true positive rates for varying classification thresholds. This curve is particularly relevant when class imbalance is severe and true positive rates are of high importance. Notably, both ROC and PR curves consider the relative ranking of predictions for binary outcome models. For multi-class models, Cohen's $\kappa$ \cite{McHugh2012} is appealing. This metric evaluates the agreement  between two labelings, accounting for agreement by chance. Finally, note that none of the presented metrics introduced above consider calibration, i.e., if model outputs reflect true likelihoods. 

To combat class imbalance, data augmentation can be used. Simple approaches include under- and oversampling (see, for instance, \cite{Lemaitre2017}). Synthetic minority oversampling (SMOTE) by Chawla \textit{et al.} \cite{Chawla2002} is another option for vector data. The technique generates convex combinations of minority class observations. Extensions include borderline-SMOTE \cite{Han2005} and borderline-SMOTE-SVM \cite{Nguyen2011}. These generate observations along estimated decision boundaries. Another SMOTE variant, ADASYN \cite{He2008}, generates observations according to data densities. For time series data (e.g., transaction sequences), there is relatively little literature on data augmentation \cite{Wen2021}. Some basic transformations are:
\begin{enumerate}
    \item window cropping, where random time series slices are extracted,
    \item window wrapping, compressing  (i.e., down-sampling),  or extending (i.e., up-sampling) time series slices,
    \item flipping, where the signs of time series are flipped (i.e., multiplied with $-1$), and
    \item noise injection, where (typically Gaussian) noise is added to time series.
\end{enumerate}
A few advanced methods also bear mentioning. Teng \textit{et al.} \cite{Teng2020} propose a wavelet transformation to preserve low-frequency time series  patterns while noise is added to high-frequency patterns.  Iwana and Uchida \cite{Iwana2021} utilize the element alignment properties of dynamic time wrapping to mix patterns; features of sample patterns are wrapped to match the time steps of reference patterns. Finally, some approaches combine multiple transformations. Cubuk \textit{et al.} \cite{Cubuk2020} propose to combine transformations at random. Fons \textit{et al.} \cite{Fons2021a} propose two adaptive schemes; the first weighs transformed observations relative to a model's loss, the second selects a subset of transformations based on rankings of prediction losses. 

Simulating known or hypothesized money laundering patterns from scratch  may be  the only option for researchers with no available data. Used together with private data sets, the approach may also ensure some reproducibility and generalizability. We refer to the work by Lopez-Rojas and Axelsson \cite{LopezRojas2012} for an in-depth discussion of simulated data for AML research. The authors develop a simulator, PaySim, for mobile phone transfers. The simulator is, in particular, employed by \cite{Buschjager2022}, proposing a generalized version of Isolation Forests to flag suspicious transactions. Weber \textit{et al.} \cite{Weber2018} and Suzumura and Kanezashi \cite{AMLSim} further augment PaySim, tailoring it to a more classic bank setting.

We have found only one public data set within the AML literature: the Elliptic data set \cite{Weber2019}. This contains a graph over bitcoin transactions. We do, however, note that graph-based approaches may be difficult to implement in a bank setting. Indeed, any bank only knows about transactions going to or from its own clients. Instead, graph approaches may be more relevant for authorities' treatment of AML reports; see work by Savage \textit{et al.} \cite{Savage2016}, Drezewski \textit{et al.} \cite{Drezewski2015}, Li \textit{et al.} \cite{Li2017}, and Baltoi \textit{et al.} \cite{Baltoiu2019}.

\subsection{Visualization, Dimension Reduction, and Semi-supervised Learning} \label{subsec:Visualization, Dimension Reduction, and Semi-supervised Techniques}
Visualization techniques may help identify money laundering \cite{Singh2019}. One option is $t$-distributed  stochastic neighbor embedding \cite{Maaten2008} and its parametric counterpart \cite{Maaten2009}. The approach is often used for 2- or 3-dimensional embeddings, aiming to keep similar observations close and dissimilar observations distant. First, a probability distribution over pairs of observations is created in the original feature space. Here, similar observations are given higher probability; dissimilar observations are given lower. Next, we seek projections that minimize the Kullback-Leibler divergence \cite{Kullback1951} to a distribution in a lower-dimensional space. Another option is ISOMAP \cite{Tenenbaum2000}. This extends multidimensional scaling \cite{Mead1992}, using the shortest path between observations to capture intrinsic similarity. 

Autoencoders, as discussed in Section \ref{subsec:Unsupervised Client Risk Profiling}, can be used both for dimension reduction, synthetic data generation, and semi-supervised learning. The latter is relevant when we have data sets with many unlabeled (but also some labeled) observations. Indeed, we may train an autoencoder with all the observations. Lower layers can then be reused in a network trained to classify labeled observations. A seminal type of autoencoder was proposed by Kingma and Welling \cite{Kingma2014}: the variational autoencoder. This is a probabilistic, generative model that seeks to minimize a loss function with two parts. The first part employs the normal reconstruction error. The second part employs the Kullback-Leibler divergence to push latent feature representations  toward  a Gaussian distribution.  An extension, conditional variational autoencoders \cite{Sohn2015} take class labels into account, modeling a conditional latent variable distribution. This allows us to generate class specific observations. Generative adversarial networks \cite{Goodfellow2014} are another option. Here, two neural networks compete against each other; a generative network produces synthetic  observations,  while a discriminative network tries to separate these from real observations. In analogy with conditional variational autoencoders, conditional generative adversarial nets \cite{Mirza2014} take class labels into account. Specifically, class labels are  fed  as inputs to both the discriminator and generator. This may, again, be used to generate class specific observations. While most generative adversarial network methods have been designed to work with visual data,  methods applicable to time-series data have recently been proposed \cite{Brophy2021,Yoon2019}.

\subsection{Neural Networks, Deep Learning, and Transfer Learning}
The neural networks used in current AML research are generally small and shallow. Deep neural networks, by contrast, employ multiple layers. The motivating idea is to derive higher-level features directly from data. This has, in particular, proved successful for computer vision \cite{Zeng2018,Hu2018}, natural language processing \cite{Radford2018, Devlin2019}, and high-frequency financial time-series analysis \cite{Tsantekidis2017, Zhang2019DeepLOB, Ntakaris2019}. Some authors have also proposed to use the approach to check KYC image information (e.g., driver's licenses) \cite{Woodruff2021,Neves2020} or ease alarm inquiries with sentiment analysis \cite{Han2020}.

State-of-the-art deep neural networks use multiple methods to combat unstable gradients. This includes rectified \cite{Nair2010,Xu2015} and exponential \cite{Clevert2016,Klambauer2017} linear units. Weight initialization is done with Xavier \cite{Glorot2010}, He \cite{He2016}, or LeCun \cite{LeCun2012} initialization. Batch normalization \cite{Ioffe2015} is used to standardize, re-scale, and shift inputs. For recurrent neural networks  (introduced below), gradient clipping \cite{Pascanu2013} and layer normalization \cite{Ba2016} are often used. Finally, residual or skip connections \cite{He2015Res,Huang2017} feed intermediate outputs multiple levels up a network hierarchy.

State-of-the-art networks also use regularization techniques to combat overfitting. Dropout \cite{Hinton2012, Srivastava2014} temporarily removes neurons during training, forcing non-dropped neurons to capture more robust relationships.  Regularization \cite{Goodfellow2016} limits  network weights by adding penalty terms to a model's loss function. Finally, max-norm regularization \cite{Srebro2005} restricts network weights directly during training. 

Multiple deep learning methods have been proposed for transfer learning. We refer to the work by Weiss \textit{et al.} \cite{Weiss2016} for an extensive review. The general idea is to utilize knowledge across different domains or tasks. One common approach starts by training a neural network on some source problem. Weights (usually from lower layers) are subsequently transferred to a new neural network that is fine-tuned (i.e., re-trained) on another target problem. This may work well when the first neural network learns to extract features that are relevant to both the source and target problem \cite{Yosinski2014}. A sub-category of transfer learning, domain adaption explicitly tries to alleviate distributional differences across domains. To this end, both unsupervised and supervised methods may be employed (depending on whether or not labeled target data is available). For example, Ganin and Lempitsky \cite{Yaroslav2015} propose an unsupervised technique that employs a gradient reversal layer and backpropagation to learn shift invariant features. Tzeng \textit{et al.} \cite{Tzeng2015} consider a semi-supervised setup where little labeled target data is available. With unlabeled target data, the authors first optimize feature representations to minimize the distance between a source and target distribution. Next, a few labeled target observations are used as reference points to adjust similarity structures among label categories. Finally, we refer to the work by Hedegaard \textit{et al.} \cite{Hedegaard2021} for a discussion and  critique  of the generic test setup used in the supervised domain adaptation literature and a proposal of a fair evaluation protocol.

Deep neural networks can, like their shallow counterparts, model sequential data. Here, we provide brief descriptions of simple instantiations of such networks. We use the notation introduced in Section \ref{subsec:Unsupervised Client Risk Profiling} and only describe single layers. To form deep learning models, one stacks multiple layers; each layer receives as input the output of its predecessor. Parameters (across all layers) are then jointly optimized by an iterative optimization scheme, as described in Section \ref{subsec:Unsupervised Client Risk Profiling}. Recurrent neural networks are one approach to  modeling  sequential data. Let $\mathbf{x}_{(t)}\in\mathbb{R}^d$ denote some layer input at time $t=1,..,T$. We can describe the time $t$ output of a basic recurrent neural network layer with $m$ neurons by
\begin{equation}
\mathbf{y}_{(t)}=\phi(\mathbf{W}_{x}^{\mathsf{T}}\mathbf{x}_{(t)}+\mathbf{W}_{y}^{\mathsf{T}}\mathbf{y}_{(t-1)}+\mathbf{b}),
\end{equation}
where $\mathbf{W}_{x}\in\mathbb{R}^{d \times m}$ is an input weight matrix, $\mathbf{W}_{y}\in\mathbb{R}^{m \times m}$ is an output weight matrix, $\mathbf{b}\in\mathbb{R}^m$ is a bias vector, and  $\phi(\cdot)$ is an activation function. Advanced architectures use gates to regulate the flow of information. Long short-term memory (LSTM) cells \cite{Hochreiter1997,sak2014long,zaremba2015recurrent} are one option. Let $\odot $ denote the Hadamard product and  $\sigma$ the standard sigmoid function. At time $t$, an LSTM layer with $m$ neurons is described by
\begin{enumerate}
    \item an input gate $\mathbf{i}_{(t)}=\sigma(\mathbf{W}_{x,i}^{\mathsf{T}}\mathbf{x}_{(t)}+\mathbf{W}_{y,i}^{\mathsf{T}}\mathbf{y}_{(t-1)}+\mathbf{b}_{i})$,
    \item a forget gate
    $\mathbf{f}_{(t)}=\sigma(\mathbf{W}_{x,f}^{\mathsf{T}}\mathbf{x}_{(t)}+\mathbf{W}_{y,f}^{\mathsf{T}}\mathbf{y}_{(t-1)}+\mathbf{b}_{f})$,
    \item an output gate
    $\mathbf{o}_{(t)}=\sigma(\mathbf{W}_{x,o}^{\mathsf{T}}\mathbf{x}_{(t)}+\mathbf{W}_{y,o}^{\mathsf{T}}\mathbf{y}_{(t-1)}+\mathbf{b}_{o})$, 
    \item a main transformation \\  \hfill ${\mathbf{g}_{(t)}=\tanh(\mathbf{W}_{x,g}^{\mathsf{T}}\mathbf{x}_{(t)}+\mathbf{W}_{y,g}^{\mathsf{T}}\mathbf{y}_{(t-1)}+\mathbf{b}_{g})}$, 
    \item a long-term state $\mathbf{l}_{(t)}=\mathbf{f}_{(t)}\odot \mathbf{l}_{(t-1)}+\mathbf{i}_{(t)} \odot  \mathbf{g}_{(t)}$, and
    \item an output $\mathbf{y}_{(t)}=\mathbf{o}_{(t)} \odot  \tanh(\mathbf{l}_{(t)})$,
\end{enumerate}
where $\mathbf{W}_{x,i},\mathbf{W}_{x,f},\mathbf{W}_{x,g},$ and $\mathbf{W}_{x,g}$ in $\mathbb{R}^{d \times m}$ denote input weight matrices, $\mathbf{W}_{y,i},\mathbf{W}_{y,f},\mathbf{W}_{y,g},$ and $\mathbf{W}_{y,g}$ in $\mathbb{R}^{m \times m}$ denote output weight matrices, and $\mathbf{b}_{i},\mathbf{b}_{f},\mathbf{b}_{o},$ and $\mathbf{b}_{g}$ in $\mathbb{R}^m$ denote biases. Cho \textit{et al.} \cite{cho2014learning} propose a simpler architecture based on gated recurrent units (called GRUs). An alternative to recurrent neural networks, the temporal neural bag-of-features architecture has proved successful for financial time series classification \cite{Passalis2020}. Here, a radial basis function layer with ${k=1,...,K}$ neurons is used
\begin{equation}
\rho(\mathbf{x}_{(t)})_{k}=\frac{\exp{(-{||}\mathbf{x}_{(t)}-\mathbf{v}_{k}\odot  \mathbf{w}_{k}{||}_{2})}}{\sum_{k=1}^{K} \exp{(-{||}\mathbf{x}_{(t)}-\mathbf{v}_{k}\odot  \mathbf{w}_{k}{||}_{2})}},
\end{equation}
where $\mathbf{v}_{k}$ and $\mathbf{w}_{k}$ in $\mathbb{R}^d$ are weights that describe the $k$'th neuron's center and width, respectively. Next, an accumulation layer is used to find a constant length representation in $\mathbb{R}^K$,
\begin{equation}
\mathbf{h}=\frac{1}{T}\sum_{t=1}^{T} 
\left(
\begin{smallmatrix}
           \rho(\mathbf{x}_{(t)})_{1} \\
           \vphantom{\int\limits^x}\smash{\vdots}\\
           \rho(\mathbf{x}_{(t)})_{K}
         \end{smallmatrix}
         \right).
\end{equation}
 Bilinear neural networks may also be used to model time domain information.  Let  ${\mathbf{X}=[\mathbf{x}_{(1)},\dots,\mathbf{x}_{(T)}]}$ be a matrix with columns $\mathbf{x}_{(t)}\in\mathbb{R}^d$ for $t=1,...,T$. A temporal bilinear layer with $m$ neurons can then be described as
\begin{equation}
\mathbf{Y}=\phi(\mathbf{W}_{1}\mathbf{X}\mathbf{W}_{2}+\mathbf{B}),
\end{equation}
where $\mathbf{W}_{1}\in\mathbb{R}^{m \times d}$ and $\mathbf{W}_{2}\in\mathbb{R}^{T \times T'}$ are weight matrices and $\mathbf{B}\in\mathbb{R}^{m \times T'}$ is a bias matrix. Notably, $\mathbf{W}_{1}$ models feature interactions at fixed time points while $\mathbf{W}_{2}$ models feature changes over time.

Attention mechanisms have recently become state-of-the-art. These allow neural networks to dynamically focus on relevant sequence elements. Bahdanau \textit{et al.} \cite{bahdanau2016neural} consider a bidirectional recurrent neural network \cite{Schuster1997} and propose a mechanism known as additive or concatenative attention. The mechanism assumes an encoder-decoder architecture. During decoding, it computes a context vector by weighing an encoder's hidden states. Weights are obtained by a secondary feedforward neural network (called an alignment model) and normalized by a softmax layer (to obtain attention scores). Notably, the secondary network is trained jointly with the primary network. Luong attention \cite{luong2015effective} is another popular mechanism, using the dot product between an encoder's and a decoder's hidden states as a similarity measure (the mechanism is also called dot-product attention). Vaswani \textit{et al.} \cite{Vaswani2017} propose the seminal transformer architecture. Here, an encoder first applies self-attention (i.e., scaled Luong attention). As before, let $\mathbf{X}=[\mathbf{x}_{(1)},\dots,\mathbf{x}_{(T)}]$ denote our matrix of sequence elements. We can describe a self-attention layer as
\begin{equation}
Z=\text{softmax}\left(\frac{\mathbf{Q}\mathbf{K}^{\mathsf{T}}}{\sqrt{d_{k}}}\right)\mathbf{V},
\end{equation}
where
\begin{enumerate}
    \item $\mathbf{Q}\in\mathbb{R}^{T\times d_{k}}$, called the query matrix, is given by ${\mathbf{Q}=\mathbf{X}^{\mathsf{T}}\mathbf{W}_{Q}}$ with a weight matrix ${\mathbf{W}_{Q}\in\mathbb{R}^{d\times d_{k}}}$,
    \item $\mathbf{K}\in\mathbb{R}^{T\times d_{k}}$, called the key matrix, is given by ${\mathbf{K}=\mathbf{X}^{\mathsf{T}}\mathbf{W}_{K}}$ with a weight matrix ${\mathbf{W}_{K}\in\mathbb{R}^{d\times d_{k}}}$, and
    \item $\mathbf{V}\in\mathbb{R}^{T\times d_{v}}$, called the value matrix, is given by ${\mathbf{V}=\mathbf{X}^{\mathsf{T}}\mathbf{W}_{V}}$ with a weight matrix ${\mathbf{W}_{V}\in\mathbb{R}^{d\times d_{v}}}$.
\end{enumerate}
Note that the softmax function is applied row-wise. It outputs a $T\times T$ matrix. Here, every row $t=1,\dots,T$ measures how much attention we pay to $\mathbf{x}_{(1)},\dots,\mathbf{x}_{(T)}$ in relation to $\mathbf{x}_{(t)}$. During decoding, the transformer also applies self-attention. Here, the key and value matrices are taken from the encoder. In addition, the decoder is only allowed to attend to earlier output sequence elements (future elements are masked, i.e., set to $-\inf$ before softmax is applied). Notably, the authors apply multiple parallel instances of self-attention. The approach, known as multi-head attention, allows attention over many abstract dimensions. Finally, positional encoding, residual connections, layer normalization, and supplementary feedforward layers are used. As a last attention mechanism, we highlight temporal attention augmented bilinear layers \cite{Tran2019}. With the notation used to introduce temporal bilinear layers above, we may express a  temporal attention augmented bilinear layer  as
\begin{enumerate}
    \item ${\mathbf{\bar{X}}}=\mathbf{W}_{1}\mathbf{X}$,
    \item $\mathbf{E}=\bar{\mathbf{X}}\mathbf{W}$,
    \item $\alpha_{(i,j)}=\frac{\exp{(e_{(i,j)})}}{\sum_{k=1}^{T}\exp{(e_{(i,k)})}},$
    \item $\mathbf{\Tilde{X}}=\lambda(\mathbf{\bar{X}}\odot \mathbf{A})+(1-\lambda)\mathbf{\bar{X}},$
    \item $\mathbf{Y}=\phi(\mathbf{\Tilde{X}}\mathbf{W}_{2})+\mathbf{B}$,
\end{enumerate}
where $a_{(i,j)}$ and $e_{(i,j)}$ denote the $(i,j)$'th element of ${\mathbf{A}\in\mathbb{R}^{m\times T}}$ and ${\mathbf{E}\in\mathbb{R}^{m\times T}}$, respectively, ${\mathbf{W}\in\mathbb{R}^{T\times T}}$ is a weight matrix with fixed diagonal elements equal to ${1/T}$, and ${\lambda\in[0,1]}$ is a scalar allowing soft attention. In particular, $\mathbf{E}$ is used to express the relative importance of temporal feature instances (learned through $\mathbf{W}$), while $\mathbf{A}$ contains our attention scores.

\subsection{Interpretable and Fair Machine Learning}
Advanced machine learning models often outperform their simple statistical counterparts. Their behavior can, however, be much harder to understand, interpret, and explain. While some supervisory authorities have shown a fair amount of leeway regarding advanced AML models \cite{Kuiper2022}, this is a potential problem. 

"Fairness" is an ambiguous concept in machine learning with many different and overlapping definitions \cite{Mehrabi2022}. The equalized odds definition states that different protected groups (e.g., genders or races) should have equal true and false positive rates. The conditional statistical parity definition takes a set of legitimate discriminative features into account, stating that the likelihood of a positive prediction should be the same across protected groups given the set of legitimate discriminative features. Finally, the counterfactual fairness definition is based on a notation that a prediction is fair if it remains unchanged in a counterfactual world where some features of interest are changed. Approaches for fair machine learning also vary greatly. In an exemplary paper, Louizos \textit{et al.} \cite{louizos2017variational} consider the use of variational autoencoders to ensure fairness. The authors treat sensitive features as nuisance or noise variables, encouraging separation between these and (informative) latent features by using factorized priors and a maximum mean discrepancy penalty term \cite{Gretton2007}. In another exemplary paper, Zhang \textit{et al.} \cite{Zhang2018} propose the use of adversarial learning. Here, a primary model tries to predict an outcome variable while minimizing an adversarial model's ability to predict protected feature values. Notably, the adversarial model takes as inputs both the primary model's predictions and other relevant features, depending on the fairness definition of interest.

Regarding interpretability, we follow Du \text{et al.} \cite{Du2019} and distinguish between intrinsic and \textit{post-hoc} interpretability. Intrinsically interpretable models are, by design, easy to understand. This includes simple decision trees and linear regression. Notably, attention mechanisms also exhibit some intrinsic interpretability; we may investigate attention scores to see what part of a particular input sequence a neural network focuses on. Other models and architectures work as ``black boxes'' and require \textit{post-hoc} interpretability methods. Here, it is useful to distinguish between global and local interpretability.  The former is concerned with overreaching model behavior; the latter with individual predictions.  One possible technique for local interpretability is 
LIME \cite{Ribeiro2016}. Consider a situation where a black box model and a single observation are given.  The method then first generates a set of permutated observations (relative to the original observation) with black box model outputs.  An intrinsically interpretable model is then trained on the synthetic data. Finally, this model is used to explain the original observation's black box prediction.  Gradient-based methods \cite{simonyan2014deep} use  the gradients associated with a particular observation and black box model to capture importance. The fundamental idea is that larger gradients (either positive or negative) imply larger feature importance. Individual conditional expectation plots \cite{Goldstein2015} are another option. These illustrate what happens to a particular black box prediction if we vary one feature value of the underlying observation. Similarly, partial dependence plots \cite{Friedman2001} may be used for global interpretability. Here, we average results from feature variations over all observations in a data set. This may, however, be misleading if input features are highly correlated. In this case, accumulated local effects plots \cite{apley2019visualizing} present an attractive alternative. These rely on conditional feature distributions and employ prediction differences. For counterfactual observation generation, numerous methods have been proposed \cite{Akula_Wang_Zhu_2020, Cheng2021, Gomez2020}. While these generally need to query an underlying model multiple times, efficient methods utilizing invertible neural networks have also been proposed \cite{hvilshoej2021ecinn}. A related problem concerns the quantitative evaluation of counterfactual examples; see the work by Hvilsh\o j \textit{et al.} \cite{hvilshoej2021quantitative} for an in-depth discussion. Finally, we highlight Shapley additive explanations (SHAP) by Lundberg and Lee \cite{Lundberg2017}. The approach is based on Shapley values \cite{Shapley1953} with a solid game-theoretical foundation. For a given observation, SHAP values  record  average marginal feature contributions (to a black box model's output) over all possible feature coalitions. The approach allows both local and global interpretability. Indeed, every observation is given a set of SHAP values (one for each input feature). Summed over the entire data set, the (numerical) SHAP values show accumulated feature importance. Although SHAP values are computationally expensive,  polynomial time estimation is possible for tree-based models \cite{Lundberg2019ExplainableAF}. 

\section{Conclusion} \label{sec:Conclusion}
Inspired by FATF’s recommendations, we propose a terminology for AML in banks structured around two central tasks: (i) client risk profiling and (ii) suspicious behavior flagging. The former assigns general risk scores to clients (e.g., for use in KYC operations) while the latter raises alarms on clients, accounts, or transactions (e.g., for use in transaction monitoring). Our review reveals that the literature on client risk profiling is characterized by diagnostics, i.e., efforts to find and explain risk factors. The literature on suspicious behavior flagging, on the other hand, is characterized by non-disclosed features and hand-crafted risk indices. 
 
In general, we find that the literature on AML in banks is plagued by a number of problems. Two challenges are class imbalance and a lack of public data sets. To address class imbalance, a multitude of different data augmentation methods may be used. Motivated by the sensitivity of bank data, synthetic data generation may be a viable way to address the lack of public data sets. Synthetic public data sets would, in particular, facilitate better evaluation and reproducibility of, as well as comparisons between, new and existing methods. Other directions for future research include methods for dimension reduction, semi-supervised learning, data visualization, deep learning, and interpretable and fair machine learning. Finally, we strongly advise against the use of accuracy as an evaluation metric for AML applications, instead emphasizing ROC or PR curves.

\end{document}